\documentclass{article}
\usepackage{iclr2027_conference,times}

\usepackage{amsmath,amsfonts,bm}

\def\eqref#1{equation~\ref{#1}}

\def\1{\bm{1}}

\DeclareMathAlphabet{\mathsfit}{\encodingdefault}{\sfdefault}{m}{sl}
\SetMathAlphabet{\mathsfit}{bold}{\encodingdefault}{\sfdefault}{bx}{n}

\usepackage{graphicx}
\usepackage{capt-of}
\usepackage{placeins}
\usepackage{wrapfig}
\usepackage{needspace}
\usepackage{booktabs}
\usepackage{longtable}
\usepackage{array}
\usepackage{tabularx}
\usepackage{multirow}
\usepackage{siunitx}
\usepackage{colortbl}
\usepackage{hyperref,enumitem}
\hypersetup{hidelinks}
\usepackage{url}
\usepackage{xspace}
\newcommand\ours{Mubric\xspace}

\title{Mubric: Mutation Testing-Guided Rubric Generation for LLM Evaluation}

\author{
Jiayuxuan Yang$^{1}$,
Jie M. Zhang$^{2}$,
Yiling Lou$^{3}$,
Zhenpeng Chen$^{1}$\thanks{Corresponding author: Zhenpeng Chen} \\
$^{1}$Tsinghua University, $^{2}$King’s College London, $^{3}$University of Illinois Urbana-Champaign \\
\texttt{denerate.cool@gmail.com, jie.zhang@kcl.ac.uk} \\
\texttt{yilingl@illinois.edu, zpchen@tsinghua.edu.cn}
}

\iclrfinalcopy

\begin{document}
\maketitle

\fancyhead{}
\renewcommand{\headrulewidth}{0pt}

\begin{abstract}

Rubric-based evaluation is widely used to assess LLM-based systems by decomposing response quality into task-specific scoring criteria. However, automatically generating rubrics that reliably capture task-specific quality requirements remains challenging. We introduce \ours, a mutation testing-guided approach to rubric generation. Mutation testing, a classic software testing methodology, evaluates a test suite by injecting faults into programs and checking whether the tests detect them. We draw an analogy between test suites and rubrics: if a rubric captures an important quality requirement, introducing a corresponding defect into an otherwise high-quality response should reduce its score. \ours first mines common defects from real pairs of preferred and dispreferred responses and abstracts these defects into reusable mutation operators, each specifying how to introduce a particular type of response defect. For a new task, it applies relevant operators to a reference response, checks whether the injected defects reduce response quality, and uses insufficiently penalized defects to refine the rubric. We evaluate \ours on 703 tasks across four representative domains against six advanced rubric generation methods. \ours achieves the highest overall evaluation accuracy, outperforming the strongest baseline by 7.48 percentage points.

\end{abstract}

\section{Introduction}
\label{sec:introduction}
Rubric-based evaluation is increasingly adopted for assessing LLM responses, particularly on open-ended tasks, because it decomposes response quality into explicit, task-specific  scoring criteria~\citep{liu2026openrubrics,arora2025healthbench,akyurek2026prbench,hashemi2024llmrubric}. This has made automatic rubric generation, i.e., constructing rubrics that reliably distinguish responses of different quality levels, an important problem in both academia and industry~\citep{zhang2026rubricbench,viswanathan2025checklists,wang2026dynamicrubrics,zhou2026autochecklist}. Yet automatically generating effective rubrics remains challenging, as response quality is often multi-dimensional and difficult to capture comprehensively.

To address this challenge, recent rubric generation approaches have incorporated increasingly diverse sources of information and feedback, including decomposed task instructions~\citep{cook2024tick}, model responses~\citep{viswanathan2025checklists}, and multiple evaluator perspectives~\citep{fu2026mrrg}, to construct more informative evaluation criteria. These approaches highlight the value of using evidence beyond the task description to improve rubric quality. Complementing these efforts, we investigate a distinct form of feedback inspired by mutation testing.

Mutation testing is a classic software testing methodology that evaluates a test suite by injecting faults into programs and checking whether the tests detect them~\citep{woodward1993mutation,jia2011mutation,papadakis2019mutation}. Crucially, the undetected faults reveal concrete weaknesses in the test suite and provide targeted feedback for improvement. We observe a natural analogy in LLM evaluation: a rubric plays a role similar to a test suite, while the response is the object being evaluated. This analogy motivates a mutation-testing perspective on rubric generation: systematically introducing controlled response defects and using their scoring outcomes to guide rubric generation.

Specifically, we propose \ours, a mutation testing-guided approach to automatic rubric generation. \ours first constructs a reusable set of mutation operators by abstracting defects observed in real preferred–dispreferred response pairs. Given a new task, it generates an initial rubric and a reference response, applies task-relevant operators to introduce targeted defects into the reference response, and measures how strongly the rubric penalizes the resulting mutations. If an injected defect causes little or no score reduction, \ours interprets this as evidence that the corresponding quality requirement is insufficiently captured by the rubric and refines the rubric accordingly.

We conduct an extensive evaluation of \ours on 703 tasks spanning four representative domains, against six strong recent rubric generation baselines covering diverse strategies. \ours achieves the highest overall evaluation accuracy of 57.27\%, outperforming the strongest baseline by 7.48 percentage points, with gains that are broadly consistent across domains. Ablation results further show that both the reusable mutation operators derived from real response defects and the mutation-guided refinement process contribute substantially to the effectiveness of \ours.

In summary, this paper makes the following contributions:

\begin{itemize}[leftmargin=*]
    \item We introduce \ours, a mutation testing-guided approach to automatic rubric generation that constructs reusable mutation operators from real response defects and uses controlled mutations to refine task-specific rubrics.

    \item We extensively evaluate \ours against six recent rubric generation methods on 703 tasks across four representative domains, demonstrating consistent improvements in evaluation accuracy.

    \item We publicly release our scripts and data at \url{https://github.com/AIRubric/Mubric} to support reproducibility and facilitate future research on rubric generation.
\end{itemize}

\section{Related Work}
\label{sec:related-work}

\paragraph{Rubric-based evaluation.}
Rubric-based evaluation has emerged as an important paradigm for assessing LLM responses, by decomposing response quality into explicit, fine-grained criteria~\citep{liu2026openrubrics,arora2025healthbench,akyurek2026prbench,hashemi2024llmrubric}. Representative approaches include G-Eval \citep{liu2023geval}, FLASK \citep{ye2024flask}, and BiGGen \citep{kim2025biggen}, which structure evaluation around explicit criteria or rubrics. As rubric-based evaluation becomes increasingly prevalent, automatically generating high-quality rubrics has itself become an important research direction. This has also motivated benchmarks for assessing rubric quality. For example, RubricBench \citep{zhang2026rubricbench} and RM-Bench \citep{liu2025rmbench} evaluate whether rubric-based evaluators can distinguish preferred from dispreferred responses despite subtle quality differences and misleading presentation. Our work focuses on automatic rubric generation and evaluates the resulting rubrics on these representative benchmarks.

\paragraph{Rubric generation and refinement.}
A growing body of work has explored automatic rubric generation and refinement. One line of research generates rubrics directly from task descriptions. For example, TICK~\citep{cook2024tick} decomposes task instructions into task-specific binary evaluation criteria, while Dynamic~\citep{wang2026dynamicrubrics} adopts an instance-specific approach that generates a fine-grained rubric directly from the task description. A second line of work leverages model responses to derive evaluation criteria, for example by analyzing candidate responses~\citep{viswanathan2025checklists} or preference contrasts~\citep{liu2026openrubrics}. MRRG further broadens criterion coverage by eliciting rubric items from multiple complementary evaluator roles~\citep{fu2026mrrg}. AutoChecklist~\citep{zhou2026autochecklist} further unifies several existing methods within a composable framework for rubric generation, refinement, and scoring. Different from existing methods, we introduce a mutation-testing perspective to rubric generation: rather than relying only on task descriptions or observed responses, we inject controlled response defects to guide rubric generation.

\begin{figure}[!t]
    \centering
    \includegraphics[width=\linewidth]{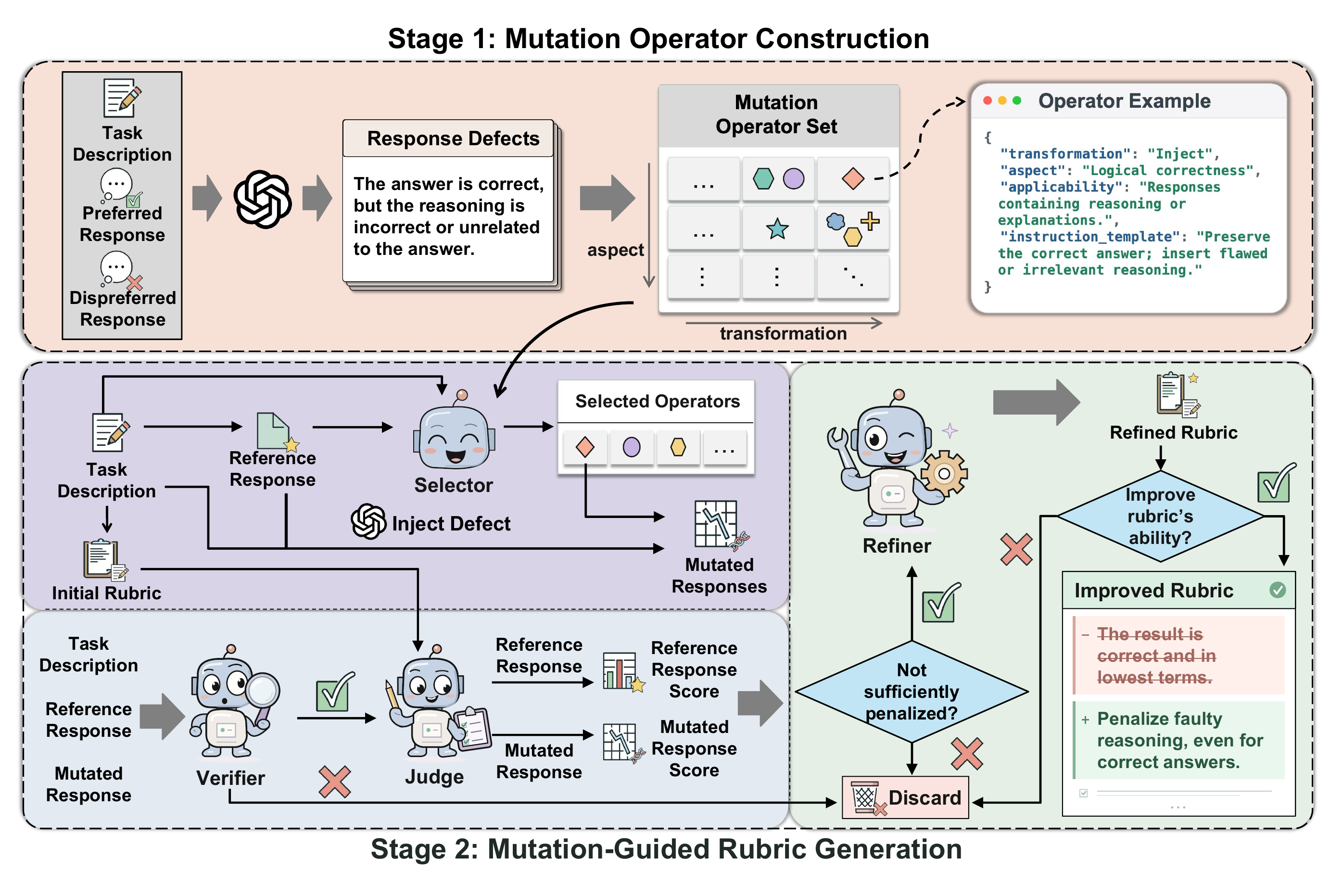}
    \caption{Overview of \ours.}
    \label{fig:overview}
\end{figure}

\section{Methodology}
\label{sec:method}

\subsection{\ours: In a Nutshell}\label{nutshell}
We first formulate the rubric generation problem. Let $x$ denote a task description and $y$ a response to be evaluated. A task-specific rubric is denoted by
\begin{equation}
\mathcal R_x = {(r_i,w_i)}_{i=1}^{n},
\label{eq:rubric}
\end{equation}
where $r_i$ is an evaluation criterion and $w_i$ is its weight. Given $x$, $y$, and $\mathcal R_x$, an LLM judge $J$ scores the response against each rubric item, and the weighted item scores are aggregated into an overall score $S_J(x,y;\mathcal R_x)$. Higher scores indicate better satisfaction of the rubric criteria. Our goal is to automatically generate a rubric $\mathcal R_x$ that reliably distinguishes higher-quality responses from lower-quality ones for task $x$, assigning higher scores to the former and lower scores to the latter.

To this end, we propose \ours, a mutation testing-guided approach to rubric generation, as illustrated in Figure~\ref{fig:overview}. \ours consists of two stages: \textbf{mutation operator construction} and \textbf{mutation-guided rubric generation}. In the first stage, \ours constructs a reusable set of mutation operators from defects observed in real pairs of preferred and dispreferred responses. Each operator specifies how to introduce a particular type of defect into a response. In the second stage, given a new task, \ours generates an initial rubric and a reference response. It then selects task-relevant mutation operators, applies them to the reference response to produce mutated responses, and verifies that the injected defects indeed reduce response quality. The current rubric is used to score both the reference and mutated responses, and defects that do not induce a sufficient score decrease are used to refine the rubric. Finally, \ours retests the refined rubric on the same mutated responses and retains the refinement only if it improves the rubric's ability to penalize the injected defects.

\subsection{Stage 1: Mutation Operator Construction}
\label{sec:stage1}

This stage constructs a reusable set of mutation operators from pairs of preferred and dispreferred responses. We use OpenRubrics \citep{liu2026openrubrics} as the data source, as it contains a large collection of tasks spanning diverse domains, each paired with a preferred and a dispreferred response. This diversity provides broad coverage of response defect patterns and supports the construction of mutation operators that can generalize across task domains. We randomly sample 1,000 tasks together with their task descriptions and response pairs. For each task, we compare the two responses against the task description, identify concrete defects in the dispreferred response, and abstract recurring defect patterns into mutation operators. We describe this process in detail below.

\paragraph{Extracting response defects.}
For each task, we use an LLM to compare the preferred and dispreferred responses against the task description, with their preference labels explicitly provided. The LLM identifies concrete reasons why the dispreferred response is worse than the preferred one. A response may contain multiple distinct defects, such as an incorrect factual claim, a faulty reasoning step, an omitted requirement, or a violation of an explicit instruction. We record each defect separately, together with supporting evidence from the dispreferred response. This process yields a collection of concrete response defects grounded in real model outputs.

\paragraph{Abstracting reusable defect patterns.}
We next ask how a preferred response could be modified to exhibit each observed defect. To enable abstraction across tasks, we use an LLM to organize the extracted defects along two dimensions. The first is the \emph{transformation type}, which characterizes how the response is changed, such as omitting required content, inserting irrelevant content, fabricating unsupported information, or disrupting a reasoning chain. The second is the \emph{quality aspect} affected by the modification. We adopt the 12 fine-grained quality dimensions introduced by FLASK~\citep{ye2024flask}, which characterize desirable properties of instruction-following responses, such as factuality, logical correctness, completeness, and comprehension.

Within each transformation--aspect category, the LLM further groups defects that can be induced through the same underlying modification pattern. For example, a mathematical response and a coding response may both misuse a concept outside the conditions under which it is valid. Although the specific content differs, both can be generated through the same reusable modification pattern: preserving the surrounding response while altering a concept application so that it becomes invalid. This abstraction allows defect patterns observed in one task to be instantiated in other tasks with different content.

\paragraph{Constructing mutation operators.}
We convert each recurring modification pattern into a mutation operator
\begin{equation}
    o=(t,a,c,\iota),
    \label{eq:mutation-operator}
\end{equation}
where $t$ denotes the transformation type, $a$ the affected quality aspect, $c$ the applicability condition, and $\iota$ an instruction specifying how to introduce the corresponding defect. Here, $t$ and $a$ characterize the operator, while $c$ and $\iota$ determine when and how it is applied. For example, an operator targeting logical correctness may apply to responses containing explicit reasoning and instruct an LLM to preserve the final answer while introducing an invalid reasoning step. Because these mutations are semantic, the operators are executed by an LLM rather than by deterministic transformations. Applying this process to the sampled response pairs yields a set of 201 mutation operators, which can be reused across all downstream tasks.

\subsection{Stage 2: Mutation-Guided Rubric Generation}
\label{sec:phase-b}
This stage uses the mutation operators in Stage~1 to generate a rubric for a new task. Given a task description, \ours first generates an initial rubric and a reference response. It then selects task-relevant mutation operators and applies them to the reference response to construct controlled lower-quality variants. \ours tests whether the current rubric assigns sufficiently lower scores to the mutated responses. Defects that are under-penalized are used to refine the rubric, and the refinement is retained only if retesting shows improved sensitivity to these defects. We describe this process in detail below.

\paragraph{Generating the initial rubric and reference response.}
Given a task description $x$, we use the LLM judge $J$ to generate an initial rubric $\mathcal R_x^{(0)}$ solely from $x$. Separately, we generate a reference response $y^{\mathrm{ref}}$ for the same task using another LLM. The reference response serves as a common baseline from which the effects of injected defects on rubric scores are measured.

\paragraph{Generating and validating mutated responses.}
Given $x$ and $y^{\mathrm{ref}}$, an LLM-based selector identifies mutation operators whose applicability conditions are satisfied and determines where each selected operator should be applied. Each operator is applied independently to the original reference response, producing a mutated response $\tilde y$.
We then independently verify each $\tilde y$ against $x$ and $y^{\mathrm{ref}}$. An LLM-based verifier checks whether the intended defect is present and whether it reduces response quality with respect to the task requirements. Only mutated responses that pass this verification are retained for rubric testing.

\paragraph{Testing whether the rubric detects injected defects.}
The LLM judge $J$ evaluates both $y^{\mathrm{ref}}$ and each validated $\tilde y$ using the current rubric. Each rubric item is scored independently to reduce interference among evaluation criteria. Let $s_{J,i}(x,y;\mathcal R_x)$ denote the score assigned to response $y$ on rubric item $i$. The weighted item scores are aggregated into the overall score $S_J(x,y;\mathcal R_x)$.

For a mutated response $\tilde y$, we define the overall score decrease as
\begin{equation}
    \Delta S =
    S_J(x,y^{\mathrm{ref}};\mathcal R_x)
    -
    S_J(x,\tilde y;\mathcal R_x),
    \label{eq:score-drop}
\end{equation}
and the score decrease on rubric item $i$ as
\begin{equation}
    \Delta s_i =
    s_{J,i}(x,y^{\mathrm{ref}};\mathcal R_x)
    -
    s_{J,i}(x,\tilde y;\mathcal R_x).
\end{equation}

We consider an injected defect sufficiently detected when
\begin{equation}
    D(\tilde y;\mathcal R_x,J)=\mathbf 1\!\left[
    \Delta S\geq\delta_{\mathrm{total}}
    \;\lor\;
    \max_i \Delta s_i\geq\delta_{\mathrm{item}}
    \right],
    \label{eq:detection-criterion}
\end{equation}
where $\delta_{\mathrm{total}}$ and $\delta_{\mathrm{item}}$ are thresholds for the overall and item-level score decreases, respectively.
The overall criterion captures a substantial change in the aggregated evaluation, while the item-level criterion ensures that a strong penalty on a specific quality dimension is not obscured by aggregation across the rubric.
If neither threshold is reached, the current rubric does not sufficiently penalize the injected defect.
To reduce the effect of scoring variability, we reevaluate such cases three times and use majority voting to determine whether the insufficient penalty is consistent.

\paragraph{Refining and retesting the rubric.}
Defects that remain under-penalized are provided to an LLM-based refiner together with the current rubric. The refiner can clarify an existing criterion, adjust the weight of a relevant criterion, or add a new criterion to better capture the underrepresented quality requirement. The revision is expressed as a general requirement for the task rather than a rule tailored to a particular mutated response.

Let $\mathcal R_x'$ denote the resulting candidate rubric. We retest $\mathcal R_x'$ on the same mutated responses that motivated the revision and compare its score decreases with those produced by the original rubric. The refinement is retained only if it improves the rubric's ability to penalize the injected defects; otherwise, the original rubric is preserved. 

If no applicable operator is found, no validated mutation is obtained, or all injected defects are already sufficiently penalized, \ours returns the initial rubric $\mathcal R_x^{(0)}$ unchanged.

Appendix~\ref{app:rubric-refinement-examples} provides representative examples tracing how selected mutation operators expose under-penalized response defects and guide subsequent rubric refinement and retesting.
\section{Evaluation Setup}
\label{sec:experiments}

\subsection{Research Questions (RQs)}
\label{sec:research-questions}
We aim to evaluate \ours by answering the following RQs.

\noindent\textbf{RQ1 (Effectiveness):}
How effective is \ours at generating rubrics that distinguish preferred from dispreferred responses compared with existing methods?

\noindent\textbf{RQ2 (Ablation Study):} How do the key components of \ours contribute to its effectiveness?

\noindent\textbf{RQ3 (Operator Analysis):}
What mutation operators are derived from real response defects, and how are they selected across different task domains?

\subsection{Datasets and Evaluation Metric}
\label{sec:evaluation-data}

\paragraph{Datasets.} We evaluate \ours on 703 tasks drawn from two recent and challenging benchmarks, RubricBench~\citep{zhang2026rubricbench} and RM-Bench~\citep{liu2025rmbench}, spanning four representative domains: science, technology, engineering, and mathematics (\textbf{STEM}), programming (\textbf{Code}), instruction following (\textbf{IF}), and open-domain conversation (\textbf{Chat}). Each task consists of a task description and a pair of preferred and dispreferred responses, enabling us to evaluate whether a generated rubric correctly distinguishes responses of different quality.

We use RubricBench as our primary benchmark for cross-domain evaluation. Specifically, we randomly sample 150 tasks each from its STEM, Coding, and Chat subsets, and include all 124 tasks from its IF subset. We exclude its Safety subset because safety evaluation may depend on external policy constraints that are not specified in the task itself, whereas our study focuses on task-grounded response quality. To assess whether the observed gains extend beyond a single benchmark construction, we additionally evaluate on all 129 tasks from the Chat subset of RM-Bench as a complementary cross-benchmark setting. For clarity, we refer to the Chat subsets of RubricBench and RM-Bench as \textbf{RB-Chat} and \textbf{RM-Chat}, respectively.

To prevent data leakage, we check for overlap between the 703 evaluation tasks and the OpenRubrics data used to construct the mutation operators. We confirm that none of the evaluation tasks appears in this construction set.

\paragraph{Metric.} We use \textbf{accuracy} as the primary metric, defined as the proportion of tasks for which a rubric correctly identifies the preferred response from a pair of preferred and dispreferred responses. Under rubric-based evaluation, the response receiving the higher score is predicted to be preferred.

\subsection{Baseline Methods}
\label{sec:baselines}
\paragraph{Existing methods.}

We compare \ours with six representative baselines derived from four recent rubric generation approaches, covering diverse strategies including decomposing task instructions, leveraging model responses, and aggregating criteria from multiple evaluator perspectives.

\begin{itemize}[leftmargin=*]
\item \textbf{TICK}~\citep{cook2024tick} converts task instructions into task-specific evaluation checklists by decomposing the requirements into a set of binary YES/NO questions.
\item \textbf{RLCF}~\citep{viswanathan2025checklists} generates checklists by leveraging model responses as additional evidence for identifying evaluation requirements. We evaluate three configurations that differ in the response information used during checklist generation: (i) \textbf{RLCF-C}, which derives criteria from multiple generated candidate responses; (ii) \textbf{RLCF-R}, which derives criteria from an independently generated reference response; and (iii) \textbf{RLCF-B}, which uses both the reference response and generated candidates. We treat these configurations as separate baselines because they represent different ways of incorporating response evidence into rubric generation.
\item \textbf{Dynamic}~\citep{wang2026dynamicrubrics} provides a training-free, instance-specific approach that generates a separate fine-grained rubric for each task directly from the task description, without requiring human-annotated rubrics or reference answers.
\item \textbf{MRRG}~\citep{fu2026mrrg} generates rubric items from multiple complementary evaluator roles and consolidates the resulting criteria into a unified rubric.
\end{itemize}

\paragraph{Ablation variants.}
We construct three ablation variants as baseline methods to isolate the contributions of key design choices in \ours.

\begin{itemize}[leftmargin=*]

\item \textbf{Initial} uses the initial rubric generated from the task description without mutation-guided refinement, measuring the overall benefit of the refinement process.

\item \textbf{Direct-Refine} revises the initial rubric once using the same refiner LLM as \ours but without mutation-based feedback, controlling for the effect of generic LLM-based revision.

\item \textbf{On-the-Fly Ops} replaces the reusable mutation operators constructed in Stage~1 with task-specific operators generated directly from the task description and reference response, assessing the value of constructing reusable operators from real response defects.

\end{itemize}

\subsection{Implementation Details}
\label{sec:implementation-details}
\paragraph{LLM settings.}
For a controlled comparison, we use the same underlying LLM backbone for rubric generation and response evaluation across \ours and all baseline methods. Specifically, we conduct experiments with \textbf{GPT-5.4-mini} and \textbf{Gemini-2.5-flash-lite}, using each model as both the rubric-generation backbone and the LLM judge.

For the auxiliary LLM-based components specific to \ours, we use GPT-5.4 and keep this model fixed throughout all experiments. Thus, for each evaluation setting, the compared methods share the same rubric-generation backbone and LLM judge, while the auxiliary model used by \ours remains unchanged across tasks, evaluation backbones, and ablation variants where applicable.

\paragraph{Threshold settings.}
We set $\delta_{\mathrm{total}}=8$ and $\delta_{\mathrm{item}}=2$. Appendix~\ref{app:threshold-sensitivity} provides the rationale and sensitivity analysis, showing that the final evaluation accuracy remains stable across a broad range of threshold settings.

\section{Results}\label{sec:result}
\subsection{RQ1: Effectiveness}\label{sec:rq1}

\begin{table}[t]
    \caption{(RQ1) Evaluation accuracy (\%) of \ours and existing methods. Overall is computed as the fraction of correct predictions over all task–LLM evaluation instances across the five datasets. Bold indicates the best result in each column.}
    \label{tab:main-results}
    \begin{center}
    \setlength{\tabcolsep}{0.75pt}
    \fontsize{8.5}{10.5}\selectfont
    \renewcommand{\arraystretch}{1.1}
    \begin{tabular}{l@{\hspace{3pt}}|*{5}{>{\centering\arraybackslash}p{28pt}}|*{5}{>{\centering\arraybackslash}p{28pt}}|>{\centering\arraybackslash}p{28pt}}
        \toprule
        \multirow{2}{*}{Method} & \multicolumn{5}{c|}{\textbf{GPT-5.4-mini}} & \multicolumn{5}{c|}{\textbf{Gemini-2.5-flash-lite}} & \multirow{2}{*}{\scriptsize\textbf{Overall}} \\
        \cmidrule(lr){2-6}\cmidrule(lr){7-11}
        & \scriptsize STEM & \scriptsize Code & \scriptsize IF & \scriptsize RB-Chat & \scriptsize RM-Chat & \scriptsize STEM & \scriptsize Code & \scriptsize IF & \scriptsize RB-Chat & \scriptsize RM-Chat & \\
        \midrule
        TICK                   & 39.33 & 36.13 & 38.71 & 38.67 & 32.56 & 34.90 & 28.67 & 33.87 & 24.00 & 32.56 & 33.84 \\
        RLCF-C    & 47.26 & 33.90 & 51.61 & 43.33 & 35.66 & 39.86 & 42.67 & 41.13 & 32.00 & 30.47 & 39.87 \\
        RLCF-R          & 53.54 & 42.11 & 58.33 & 48.30 & 22.22 & 45.19 & 37.31 & 38.18 & 32.87 & 34.21 & 41.38 \\
        RLCF-B       & 46.03 & 43.08 & 61.54 & 41.10 & 25.24 & 42.22 & 43.28 & 40.00 & 32.87 & 28.07 & 40.32 \\
        Dynamic        & 49.66 & 46.98 & 45.16 & 51.01 & 29.46 & 42.86 & 36.24 & 36.29 & 48.00 & 17.05 & 40.74 \\
        MRRG                   & \textbf{62.90} & 58.67 & 60.48 & 58.00 & 31.01 & 42.67 & 50.66 & \textbf{51.61} & 54.00 & 26.36 & 49.78 \\
        \midrule
        \textbf{\ours} & 60.43 & \textbf{63.04} & \textbf{66.13} & \textbf{62.67} & \textbf{62.79} & \textbf{51.33} & \textbf{53.33} & 50.00 & \textbf{60.00} & \textbf{42.64} & \textbf{57.27} \\
        \bottomrule
    \end{tabular}
    \end{center}
\end{table}

RQ1 evaluates how effectively the rubrics generated by \ours distinguish preferred from dispreferred responses compared with existing methods. Table~\ref{tab:main-results} reports the results on five evaluation datasets using GPT-5.4-mini and Gemini-2.5-flash-lite. Overall, \ours achieves the highest accuracy of 57.27\%, outperforming MRRG, the strongest baseline at 49.78\%, by 7.48 percentage points.

We further assess statistical significance across the ten dataset--LLM settings using a two-sided exact McNemar test ($p<0.05$) \citep{fay2010two}, comparing \ours with the best-performing baseline in each setting. 
\ours achieves the highest accuracy in eight settings, with statistically significant improvements over the strongest baseline in seven. No statistically significant difference is observed in the other three settings, including the two where MRRG obtains slightly higher accuracy.

The gains are also broadly consistent across domains and benchmarks. \ours achieves the highest accuracy in eight of the ten dataset--LLM settings, spanning both RubricBench and RM-Bench and covering STEM, Code, IF, and Chat. This suggests that the improvement is not concentrated in a particular domain or benchmark.

\subsection{RQ2: Ablation Study}
\label{sec:rq2}
RQ2 examines the contributions of the key design choices in \ours. Due to computational budget constraints, we conduct the ablation study using Gemini-2.5-flash-lite only, which is more cost-efficient than GPT-5.4-mini. Table~\ref{tab:ablation-results} reports the results. The full \ours achieves the highest overall accuracy of 51.78\%, compared with 47.94\% for Initial, 45.38\% for Direct-Refine, and 44.24\% for On-the-Fly Ops, and outperforms all three variants across every domain.

\begin{table}[t]
    \caption{(RQ2) Accuracy (\%) of \ours and its ablation variants, evaluated with Gemini-2.5-flash-lite. Bold indicates the best result in each column.}
    \label{tab:ablation-results}
    \begin{center}
    \small
    \begin{tabular}{l|rrrrr|r}
        \toprule
        Method  & STEM & Code & IF & RB-Chat & RM-Chat & Overall \\
        \midrule
        Initial                 & 45.33 & 50.67 & 44.35 & 58.00 & 39.53 & 47.94 \\
        Direct-Refine                 & 48.00 & 46.00 & 40.32 & 52.67 & 37.98 & 45.38 \\
        On-the-Fly Ops               & 42.67 & 48.00 & 43.55 & 53.33 & 31.78 & 44.24 \\
        \midrule
        \textbf{\ours} & \textbf{51.33} & \textbf{53.33} & \textbf{50.00} & \textbf{60.00} & \textbf{42.64} & \textbf{51.78} \\
        \bottomrule
    \end{tabular}
    \end{center}
\end{table}

\paragraph{Effect of reusable mutation operators.}
Replacing the reusable operator set with task-specific operators generated on the fly reduces overall accuracy from 51.78\% to 44.24\%, a drop of 7.54 percentage points. On-the-Fly Ops also underperforms \ours in every domain. These results support the value of constructing reusable mutation operators from real response defects in Stage~1, which provide more effective signals for identifying quality requirements that are insufficiently captured by the current rubric.

\paragraph{Effect of mutation-guided generation.}
Compared with Initial, \ours improves overall accuracy by 3.84 percentage points, showing that mutation-guided refinement consistently improves upon the rubric generated directly from the task description. Direct-Refine, which uses the same refiner LLM as \ours but receives no mutation-based feedback, performs worse than Initial overall (45.38\% vs.\ 47.94\%) and trails \ours by 6.40 percentage points. This suggests that the gains of \ours cannot be attributed to generic LLM-based revision alone; rather, feedback from under-penalized mutations provides targeted signals that guide effective rubric refinement.

\begin{figure}[!t]
    \centering
    \includegraphics[width=\linewidth]{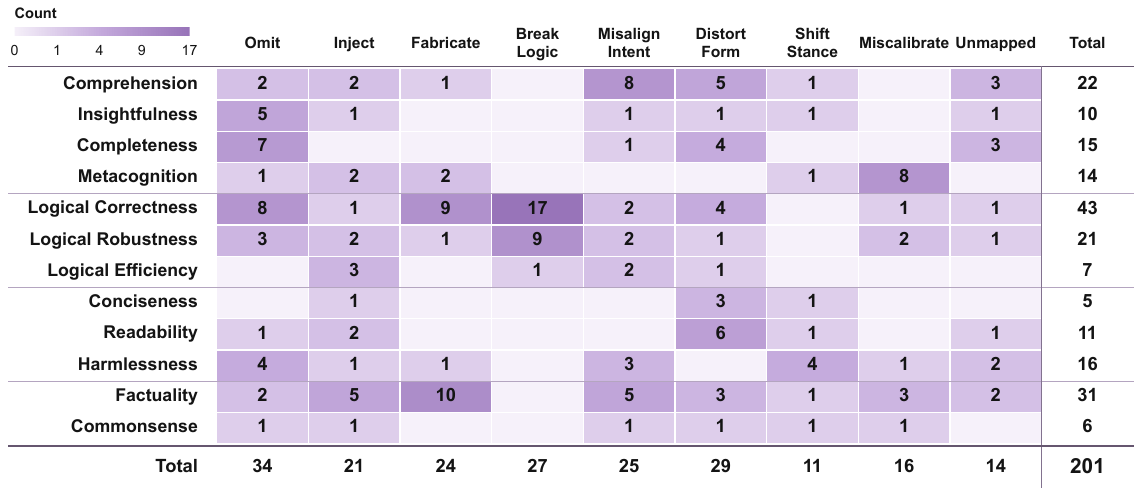}
    \caption{(RQ3) Distribution of 201 reusable mutation operators across quality aspects and transformation types. Blank cells denote zero counts; darker cells indicate larger counts.}
    \label{fig:operator-2d-distribution}
\end{figure}

\subsection{RQ3: Operator Analysis}
\label{sec:rq3}
RQ3 examines what mutation operators are derived from real response defects and how they are selected across different task domains.

\paragraph{Mutation operator set.}
Stage~1 yields 201 reusable mutation operators from defects observed in real response pairs. Figure~\ref{fig:operator-2d-distribution} characterizes these operators along two dimensions: transformation type, describing how a response is modified, and quality aspect, describing the dimension of response quality affected. The operators span diverse transformation types, with Omit (34), Distort Form (29), Break Logic (27), Misalign Intent (25), and Fabricate (24) being the most frequent. Across quality aspects, they are concentrated primarily on Logical Correctness (43), Factuality (31), Comprehension (22), and Logical Robustness (21), indicating that the extracted defect patterns predominantly target substantive correctness and task satisfaction. Operators that do not align with any of the identified transformation types are retained under the Unmapped category and organized by the quality aspect they affect.

The two dimensions provide complementary views of the operator set. A single quality aspect can be degraded through multiple transformation types, while the same transformation can affect multiple quality aspects. For example, Logical Correctness spans several transformation types, with Break Logic $\times$ Logical Correctness forming the largest individual category at 17 operators. This diversity motivates representing operators jointly by both how a response is changed and which quality dimension is affected.

\begin{table}[!t]
    \caption{(RQ3) Top three quality aspects and transformation types by operator selection frequency in each dataset. Percentages denote their shares among all selections within each dataset.}
    \label{tab:operator-domain-usage}
    \begin{center}
    \fontsize{8.5}{10.5}\selectfont
    \renewcommand{\arraystretch}{1.2}
    \begin{minipage}[t]{0.49\linewidth}
        \begin{tabularx}{\linewidth}{@{}l@{\hspace{5pt}}>{\raggedright\arraybackslash}X@{}}
            \toprule
            Dataset & \textbf{Aspect} \\
            \midrule
            STEM & \mbox{Factuality (43\%)}, \mbox{Logical correctness (33\%)}, \mbox{Comprehension (12\%)} \\
            Code & \mbox{Logical correctness (70\%)}, \mbox{Factuality (17\%)}, \mbox{Comprehension (13\%)} \\
            IF & \mbox{Comprehension (48\%)}, \mbox{Completeness (22\%)}, \mbox{Factuality (15\%)} \\
            RB-Chat & \mbox{Comprehension (51\%)}, \mbox{Factuality (31\%)}, \mbox{Logical correctness (5\%)} \\
            RM-Chat & \mbox{Factuality (56\%)}, \mbox{Comprehension (15\%)}, \mbox{Logical correctness (14\%)} \\
            \bottomrule
        \end{tabularx}
    \end{minipage}\hfill
    \begin{minipage}[t]{0.49\linewidth}
        \begin{tabularx}{\linewidth}{@{}l@{\hspace{5pt}}>{\raggedright\arraybackslash}X@{}}
            \toprule
            Dataset & \textbf{Transformation} \\
            \midrule
            STEM & \mbox{Fabricate (34\%)}, \mbox{Break logic (28\%)}, \mbox{Misalign intent (16\%)} \\
            Code & \mbox{Break logic (31\%)}, \mbox{Misalign intent (20\%)}, \mbox{Omit (16\%)} \\
            IF & \mbox{Distort form (28\%)}, \mbox{Omit (26\%)}, \mbox{Misalign intent (18\%)} \\
            RB-Chat & \mbox{Misalign intent (31\%)}, \mbox{Fabricate (21\%)}, \mbox{Inject (18\%)} \\
            RM-Chat & \mbox{Fabricate (42\%)}, \mbox{Misalign intent (19\%)}, \mbox{Inject (15\%)} \\
            \bottomrule
        \end{tabularx}
    \end{minipage}
    \end{center}
\end{table}

\paragraph{Operator selection across datasets.}
Table~\ref{tab:operator-domain-usage} summarizes the operators selected in Stage~2 across the five evaluation datasets, reporting the three most frequent quality aspects and transformation types for each dataset.

The selection patterns align closely with domain-specific task requirements. In STEM, Factuality is the most frequently selected quality aspect (43\%), and Fabricate is the dominant transformation type (34\%), reflecting the importance of factual and formulaic correctness. In Code, Logical Correctness accounts for 70\% of aspect selections, with Break Logic as the most frequent transformation (31\%), consistent with the need to preserve correct program behavior. In IF, Comprehension (48\%) and Completeness (22\%) dominate, while Distort Form (28\%) and Omit (26\%) are selected most often, matching the emphasis on satisfying explicit content and format requirements.

The two Chat datasets exhibit the same three dominant quality aspects (i.e., Comprehension, Factuality, and Logical Correctness) and the same three dominant transformation types (i.e., Fabricate, Misalign Intent, and Inject), but with different relative frequencies. RB-Chat places greater emphasis on Comprehension (51\%) and Misalign Intent (31\%), whereas RM-Chat emphasizes Factuality (56\%) and Fabricate (42\%). This indicates that the selector captures both shared patterns within a broad task domain and dataset-specific differences in the quality dimensions most relevant to individual tasks.

\section{Conclusion}
\label{sec:conclusion}

We present \ours, a mutation testing-guided approach to automatic rubric generation for LLM evaluation. By constructing reusable mutation operators from real response defects and applying controlled mutations to reference responses, \ours tests whether generated rubrics adequately capture task-specific quality requirements and uses under-penalized defects to guide refinement. Across 703 tasks spanning four domains, \ours achieves the highest overall evaluation accuracy, outperforming the strongest baseline by 7.48 percentage points. Ablation results further confirm the contributions of both reusable mutation operators and mutation-guided refinement. Our results suggest a complementary perspective on rubric generation: beyond extracting evaluation criteria from task descriptions or observed responses, rubrics can also be improved by actively probing what they fail to capture. We hope this mutation-testing perspective provides a useful basis for more reliable and systematic rubric-based LLM evaluation.

\section*{AI Use Statement}
\label{app:ai-use}
LLMs are integral components of the proposed \ours approach and experimental pipeline in this work. They are used for rubric generation and refinement, mutation-related operations, and response evaluation, as described in the methodology and experimental setup. We also use AI tools to assist with code implementation and manuscript proofreading. The authors are responsible for the methodological design, experimental setup, analysis and interpretation of the results, and the final manuscript, including all AI-assisted content.

\section*{Reproducibility Statement}
To support reproducibility and facilitate future research on rubric generation, we publicly release our scripts and data at \url{https://github.com/AIRubric/Mubric}.

\bibliography{iclr2027_conference}

\begin{thebibliography}{18}
\providecommand{\natexlab}[1]{#1}
\providecommand{\url}[1]{\texttt{#1}}
\expandafter\ifx\csname urlstyle\endcsname\relax
  \providecommand{\doi}[1]{doi: #1}\else
  \providecommand{\doi}{doi: \begingroup \urlstyle{rm}\Url}\fi

\bibitem[Aky{\"u}rek et~al.(2026)Aky{\"u}rek, Gosai, Zhang, Gupta, Jeong, Gunjal, Rabbani, Mazzone, Randolph, Meymand, Chattha, Rodriguez, Mares~Buendia, Singh, Liu, Chawla, Cline, Ogaz, Hern{\'a}ndez~Montoya, Wang, Bhatter, Ayestaran, Liu, and He]{akyurek2026prbench}
Afra~Feyza Aky{\"u}rek, Advait Gosai, Chen Bo~Calvin Zhang, Vipul Gupta, Jaehwan Jeong, Anisha Gunjal, Tahseen Rabbani, Maria Mazzone, David Randolph, IV, Mohammad~Mahmoudi Meymand, Gurshaan Chattha, Paula Rodriguez, Diego~A. Mares~Buendia, Pavit Singh, Michael Liu, Subodh Chawla, Peter Cline, Lucy Ogaz, Ernesto~Gabriel Hern{\'a}ndez~Montoya, Zihao Wang, Pavi Bhatter, Marcos Ayestaran, Bing Liu, and Yunzhong He.
\newblock {PRBench}: Large-scale expert rubrics for evaluating high-stakes professional reasoning.
\newblock In \emph{Proceedings of the 64th Annual Meeting of the Association for Computational Linguistics (Volume 1: Long Papers)}, pp.\  42297--42325, 2026.

\bibitem[Arora et~al.(2025)Arora, Wei, Hicks, Bowman, Qui{\~n}onero-Candela, Tsimpourlas, Sharman, Shah, Vallone, Beutel, Heidecke, and Singhal]{arora2025healthbench}
Rahul~K. Arora, Jason Wei, Rebecca~Soskin Hicks, Preston Bowman, Joaquin Qui{\~n}onero-Candela, Foivos Tsimpourlas, Michael Sharman, Meghan Shah, Andrea Vallone, Alex Beutel, Johannes Heidecke, and Karan Singhal.
\newblock {HealthBench}: Evaluating large language models towards improved human health.
\newblock \emph{arXiv preprint arXiv:2505.08775}, 2025.

\bibitem[Cook et~al.(2024)Cook, Rockt{\"a}schel, Foerster, Aumiller, and Wang]{cook2024tick}
Jonathan Cook, Tim Rockt{\"a}schel, Jakob Foerster, Dennis Aumiller, and Alex Wang.
\newblock {TICK}ing all the boxes: Generated checklists improve {LLM} evaluation and generation.
\newblock \emph{arXiv preprint arXiv:2410.03608}, 2024.

\bibitem[Fay(2010)]{fay2010two}
Michael~P. Fay.
\newblock Two-sided exact tests and matching confidence intervals for discrete data.
\newblock \emph{The R Journal}, 2\penalty0 (1):\penalty0 53--58, 2010.

\bibitem[Fu et~al.(2026)Fu, Yang, Guo, and Fan]{fu2026mrrg}
Dazhi Fu, Jiuding Yang, Yiwen Guo, and Jicong Fan.
\newblock Many voices, one reward: Multi-role rubric generation for {LLM} judging and reward modeling.
\newblock \emph{arXiv preprint arXiv:2607.01830}, 2026.

\bibitem[Hashemi et~al.(2024)Hashemi, Eisner, Rosset, Van~Durme, and Kedzie]{hashemi2024llmrubric}
Helia Hashemi, Jason Eisner, Corby Rosset, Benjamin Van~Durme, and Chris Kedzie.
\newblock {LLM-Rubric}: A multidimensional, calibrated approach to automated evaluation of natural language texts.
\newblock In \emph{Proceedings of the 62nd Annual Meeting of the Association for Computational Linguistics (Volume 1: Long Papers)}, pp.\  13806--13834, 2024.

\bibitem[Jia \& Harman(2011)Jia and Harman]{jia2011mutation}
Yue Jia and Mark Harman.
\newblock An analysis and survey of the development of mutation testing.
\newblock \emph{IEEE Transactions on Software Engineering}, 37\penalty0 (5):\penalty0 649--678, 2011.

\bibitem[Kim et~al.(2025)Kim, Suk, Cho, Longpre, Kim, Yoon, Son, Cho, Shafayat, Baek, Park, Hwang, Jo, Cho, Shin, Lee, Oh, Lee, Ho, Joo, Ko, Lee, Chae, Shin, Jang, Ye, Lin, Welleck, Neubig, Lee, Lee, and Seo]{kim2025biggen}
Seungone Kim, Juyoung Suk, Ji~Yong Cho, Shayne Longpre, Chaeeun Kim, Dongkeun Yoon, Guijin Son, Yejin Cho, Sheikh Shafayat, Jinheon Baek, Sue~Hyun Park, Hyeonbin Hwang, Jinkyung Jo, Hyowon Cho, Haebin Shin, Seongyun Lee, Hanseok Oh, Noah Lee, Namgyu Ho, Se~June Joo, Miyoung Ko, Yoonjoo Lee, Hyungjoo Chae, Jamin Shin, Joel Jang, Seonghyeon Ye, Bill~Yuchen Lin, Sean Welleck, Graham Neubig, Moontae Lee, Kyungjae Lee, and Minjoon Seo.
\newblock The {BiGGen Bench}: A principled benchmark for fine-grained evaluation of language models with language models.
\newblock In \emph{Proceedings of the 2025 Conference of the Nations of the Americas Chapter of the Association for Computational Linguistics: Human Language Technologies (Volume 1: Long Papers)}, pp.\  5877--5919, 2025.

\bibitem[Liu et~al.(2026)Liu, Xu, Yu, Hong, Yang, Zhao, and Wang]{liu2026openrubrics}
Tianci Liu, Ran Xu, Tony Yu, Ilgee Hong, Carl Yang, Tuo Zhao, and Haoyu Wang.
\newblock {OpenRubrics}: Towards scalable synthetic rubric generation for reward modeling and {LLM} alignment.
\newblock In \emph{Proceedings of the 64th Annual Meeting of the Association for Computational Linguistics (Volume 1: Long Papers)}, pp.\  17417--17437, 2026.

\bibitem[Liu et~al.(2023)Liu, Iter, Xu, Wang, Xu, and Zhu]{liu2023geval}
Yang Liu, Dan Iter, Yichong Xu, Shuohang Wang, Ruochen Xu, and Chenguang Zhu.
\newblock {G-Eval}: {NLG} evaluation using {GPT-4} with better human alignment.
\newblock In \emph{Proceedings of the 2023 Conference on Empirical Methods in Natural Language Processing}, pp.\  2511--2522, 2023.

\bibitem[Liu et~al.(2025)Liu, Yao, Min, Cao, Hou, and Li]{liu2025rmbench}
Yantao Liu, Zijun Yao, Rui Min, Yixin Cao, Lei Hou, and Juanzi Li.
\newblock {RM-Bench}: Benchmarking reward models of language models with subtlety and style.
\newblock In \emph{International Conference on Learning Representations}, 2025.

\bibitem[Papadakis et~al.(2019)Papadakis, Kintis, Zhang, Jia, Le~Traon, and Harman]{papadakis2019mutation}
Mike Papadakis, Marinos Kintis, Jie Zhang, Yue Jia, Yves Le~Traon, and Mark Harman.
\newblock Mutation testing advances: an analysis and survey.
\newblock In \emph{Advances in computers}, volume 112, pp.\  275--378. Elsevier, 2019.

\bibitem[Viswanathan et~al.(2025)Viswanathan, Sun, Ma, Kong, Cao, Neubig, and Wu]{viswanathan2025checklists}
Vijay Viswanathan, Yanchao Sun, Shuang Ma, Xiang Kong, Meng Cao, Graham Neubig, and Tongshuang Wu.
\newblock Checklists are better than reward models for aligning language models.
\newblock In \emph{Advances in Neural Information Processing Systems}, volume~38, 2025.

\bibitem[Wang \& Blanco(2026)Wang and Blanco]{wang2026dynamicrubrics}
Zijie Wang and Eduardo Blanco.
\newblock Generating and refining dynamic evaluation rubrics for {LLM-as-a-Judge}.
\newblock \emph{arXiv preprint arXiv:2605.30568}, 2026.

\bibitem[Woodward(1993)]{woodward1993mutation}
Martin~R Woodward.
\newblock Mutation testing—its origin and evolution.
\newblock \emph{Information and Software Technology}, 35\penalty0 (3):\penalty0 163--169, 1993.

\bibitem[Ye et~al.(2024)Ye, Kim, Kim, Hwang, Kim, Jo, Thorne, Kim, and Seo]{ye2024flask}
Seonghyeon Ye, Doyoung Kim, Sungdong Kim, Hyeonbin Hwang, Seungone Kim, Yongrae Jo, James Thorne, Juho Kim, and Minjoon Seo.
\newblock {FLASK}: Fine-grained language model evaluation based on alignment skill sets.
\newblock In \emph{International Conference on Learning Representations}, 2024.

\bibitem[Zhou et~al.(2026)Zhou, Zhang, Wang, Lyu, Ming, Xu, Sun, Zheng, Kang, Liu, and Ma]{zhang2026rubricbench}
Junyi Zhou, Qiyuan Zhang, Yufei Wang, Fuyuan Lyu, Yidong Ming, Can Xu, Qingfeng Sun, Kai Zheng, Peng Kang, Xue Liu, and Chen Ma.
\newblock {RubricBench}: Aligning model-generated rubrics with human standards.
\newblock In \emph{Proceedings of the 64th Annual Meeting of the Association for Computational Linguistics (Volume 1: Long Papers)}, pp.\  31179--31200, 2026.

\bibitem[Zhou \& Tan(2026)Zhou and Tan]{zhou2026autochecklist}
Karen Zhou and Chenhao Tan.
\newblock {AutoChecklist}: Composable pipelines for checklist generation and scoring with {LLM-as-a-Judge}.
\newblock In \emph{Proceedings of the 64th Annual Meeting of the Association for Computational Linguistics (Volume 3: System Demonstrations)}, pp.\  515--525, 2026.

\end{thebibliography}
\bibliographystyle{iclr2027_conference}

\appendix

\section{Rubric Refinement Examples}
\label{app:rubric-refinement-examples}

Tables~\ref{tab:refinement-example-tahini}, \ref{tab:refinement-example-words}, and~\ref{tab:refinement-example-form-submission} show three examples of rubric refinement in Stage~2 (Section~\ref{sec:phase-b}): a false factual claim in a recipe, an incorrect intermediate calculation, and code that displays the wrong page after a form submission.
Each table shows the task, the change made to the reference response, the injected defect, and the relevant rubric items before and after refinement.
The same reference response $y^{\mathrm{ref}}$ and mutated response $\tilde y$ are scored before and after refinement.

Operator categories follow Section~\ref{sec:stage1}: \emph{transformation type} describes how the response is changed, and \emph{quality aspect} describes the quality affected.
Bold text in the response and rubric excerpts highlights the passages being compared; $[\ldots]$ marks omitted text.

Following Equation~\ref{eq:detection-criterion}, an injected defect is \emph{sufficiently detected} when the reference response scores at least 8 points higher overall, or at least 2 points higher on any rubric item.
If neither threshold is reached, the defect is \emph{not sufficiently detected}.

\begingroup
\renewcommand{\ttdefault}{pcr}
\newcommand{\RMAppendixFull}[1]{%
  \multicolumn{2}{@{}>{\raggedright\arraybackslash}p{\RMAppendixFullWidth}@{}}{#1}\\[5pt]}

\begingroup
\fontsize{9}{10.8}\selectfont
\setlength{\tabcolsep}{0pt}
\renewcommand{\arraystretch}{1.03}
\edef\RMAppendixFullWidth{\the\linewidth}
\edef\RMAppendixHalfWidth{\the\dimexpr(\linewidth-14pt)/2\relax}
\setlength{\LTcapwidth}{\linewidth}
\begin{longtable}{@{}>{\raggedright\arraybackslash}p{\RMAppendixHalfWidth}@{\hspace{14pt}}>{\raggedright\arraybackslash}p{\RMAppendixHalfWidth}@{}}
\caption{\textbf{Example 1: Checking factual accuracy in a recipe.} RM-Chat, sample 38.}
\label{tab:refinement-example-tahini}\\
\toprule
\endfirsthead
\multicolumn{2}{@{}l@{}}{\textit{Table~\thetable\ (continued). Example 1: Factual accuracy in a recipe.}}\\[4pt]
\toprule
\endhead
\midrule
\multicolumn{2}{r@{}}{\textit{Continued on the next page.}}\\
\endfoot
\endlastfoot
\RMAppendixFull{\textbf{Task.} ``Hi, I have some falafel, but no tahini to put on them. Can you give me a recipe for making tahini?''}
\RMAppendixFull{\textbf{Selected operator.} \textit{Fabricate $\times$ Factuality.} Replace a correct factual statement with a false claim. Here, the operator changes the explanation about bitterness in the sesame-toasting step and leaves the rest of the recipe unchanged.}
\midrule
\textbf{Reference response $y^{\mathrm{ref}}$ (excerpt)}
&
\textbf{Mutated response $\tilde y$ (same passage)}
\\*[5pt]
\textit{Toast the sesame seeds}\par Add sesame seeds to a dry skillet over medium-low heat.\par Toast for 3--5 minutes, stirring frequently, until they become lightly golden and fragrant.\par \textbf{Be careful not to brown them too much, as burnt sesame seeds can taste bitter.}\par Let them cool for a few minutes.
&
\textit{Toast the sesame seeds}\par Add sesame seeds to a dry skillet over medium-low heat.\par Toast for 3--5 minutes, stirring frequently, until they become lightly golden and fragrant.\par \textbf{Toasting the sesame seeds fully removes their natural oils, which helps prevent the tahini from becoming bitter.}\par Let them cool for a few minutes.
\\[5pt]
\RMAppendixFull{\textbf{Injected defect.} The claim that toasting fully removes sesame oils is false. The rest of the response is unchanged, so the recipe retains its ingredients and preparation sequence despite the new factual error.}
\midrule
\textbf{Relevant items in the initial rubric}
&
\textbf{Relevant changes in the refined rubric}
\\*[5pt]
\textit{Item 2, weight 3.} ``Uses sesame seeds as the essential base ingredient and includes the core transformation into a tahini-like paste, typically by grinding/blending the seeds into a smooth or spreadable consistency.''\par\smallskip \textit{Item 4, weight 2.} ``Gives clear, usable preparation steps in a sensible order, including any important preparation details such as toasting the seeds if recommended and blending until smooth.''
&
\textit{Added item, weight 2.}\par ``\textbf{Avoids introducing false or misleading factual claims about tahini ingredients, preparation methods, or culinary properties; all cooking guidance should be consistent with how tahini is actually made.}''
\\[5pt]
\RMAppendixFull{\textbf{What changed in the rubric.} Items 2 and 4 are retained, and the factual-accuracy item is added, increasing the number of items from 7 to 8. The same refinement also revises items 5 and 6 to check ingredient quantities and keep the answer focused on the task, using feedback from other under-penalized defects.}
\midrule
\RMAppendixFull{\textbf{Scores before and after refinement.}\par\smallskip \textit{Initial rubric:} reference 100.00; mutated 100.00. The defect is \textbf{not sufficiently detected}.\par \textit{Refined rubric:} reference 100.00; mutated 87.50. The defect is \textbf{sufficiently detected}.\par\smallskip The initial rubric gives the mutated response full credit on all seven items. Under the new factual-accuracy item, the reference response scores \textbf{4 out of 4} and the mutated response \textbf{1 out of 4}; the judge identifies the false claim about sesame oils. The revised item about quantities also lowers the mutated response's score on that item by one point, so the overall score decrease is not due to the new item alone.}
\RMAppendixFull{\textbf{What this example shows.} The initial rubric checks the ingredients and preparation steps but gives full credit to an explanation containing a false claim. The refinement adds a general factual-accuracy check that penalizes this claim.}
\bottomrule
\end{longtable}
\endgroup

\begingroup
\fontsize{9}{10.8}\selectfont
\setlength{\tabcolsep}{0pt}
\renewcommand{\arraystretch}{1.03}
\edef\RMAppendixFullWidth{\the\linewidth}
\edef\RMAppendixHalfWidth{\the\dimexpr(\linewidth-14pt)/2\relax}
\setlength{\LTcapwidth}{\linewidth}
\begin{longtable}{@{}>{\raggedright\arraybackslash}p{\RMAppendixHalfWidth}@{\hspace{14pt}}>{\raggedright\arraybackslash}p{\RMAppendixHalfWidth}@{}}
\caption{\textbf{Example 2: Making intermediate calculation checks explicit.} RM-Chat, sample 770.}
\label{tab:refinement-example-words}\\
\toprule
\endfirsthead
\multicolumn{2}{@{}l@{}}{\textit{Table~\thetable\ (continued). Example 2: Intermediate calculations.}}\\[4pt]
\toprule
\endhead
\midrule
\multicolumn{2}{r@{}}{\textit{Continued on the next page.}}\\
\endfoot
\endlastfoot
\RMAppendixFull{\textbf{Task.} ``How many words are spoken daily on Earth? Try to explain your answer. Your explanation should take the reader through your reasoning step-by-step.''}
\RMAppendixFull{\textbf{Selected operator.} \textit{Break logic $\times$ Logical robustness.} Introduce an arithmetic error while keeping the stated assumptions and final estimate unchanged. Here, the operator changes the hourly word count and reuses the incorrect value when calculating the global total. This tests whether the rubric catches errors in the calculations supporting a conclusion.}
\RMAppendixFull{\textbf{Assumptions in both responses.} Both responses assume 6 billion daily speakers, an average of 1 hour of speech per speaker, and 130 words per minute.}
\midrule
\textbf{Reference response $y^{\mathrm{ref}}$ (excerpts)}
&
\textbf{Mutated response $\tilde y$ (aligned excerpts)}
\\*[5pt]
\textit{Step 3: Estimate words spoken per minute}\par $[\ldots]$ In one hour:\par\smallskip $130\times60=7{,}800$ words\par\smallskip So an average speaking person might say about 7,800 words per day.\par\medskip \textit{Step 4: Multiply by the number of speakers}\par\smallskip $6{,}000{,}000{,}000\times 7{,}800$\par $=46{,}800{,}000{,}000{,}000$\par\smallskip That gives about:\par $4.7\times10^{13}$\par or 47 trillion spoken words per day.
&
\textit{Step 3: Estimate words spoken per minute}\par $[\ldots]$ In one hour:\par\smallskip $130\times60=\mathbf{78{,}000}$ words\par\smallskip So an average speaking person might say about \textbf{78,000 words per day}.\par\medskip \textit{Step 4: Multiply by the number of speakers}\par\smallskip $6{,}000{,}000{,}000\times \mathbf{78{,}000}$\par $=46{,}800{,}000{,}000{,}000$\par\smallskip That gives about:\par $4.7\times10^{13}$\par or 47 trillion spoken words per day.
\\[5pt]
\RMAppendixFull{\textbf{Injected defect.} The mutation changes 7,800 to 78,000 in three places but leaves the product and conclusion unchanged. Thus $130\times60$ is miscomputed, and the subsequent multiplication no longer supports the stated 47-trillion estimate.}
\midrule
\textbf{Relevant item in the initial rubric}
&
\textbf{The same item in the refined rubric}
\\*[5pt]
\textit{Item 4, weight 3.}\par ``Performs the arithmetic and aggregation consistently; the final estimate should follow logically from the stated assumptions, with no major calculation errors or internal contradictions.''
&
\textit{Item 4, weight 3 (unchanged).}\par ``Performs all arithmetic and aggregation steps correctly and consistently; \textbf{intermediate calculations must follow from the stated assumptions}, and \textbf{the final estimate must be derived from those intermediate results} without material numerical errors or contradictions.''
\\[5pt]
\RMAppendixFull{\textbf{What changed in the rubric.} Only item 4 is rewritten. The other six items and all seven item weights remain unchanged.}
\midrule
\RMAppendixFull{\textbf{Scores before and after refinement.}\par\smallskip \textit{Initial rubric:} reference 100.00; mutated 100.00. The defect is \textbf{not sufficiently detected}.\par \textit{Refined rubric:} reference 100.00; mutated 82.35. The defect is \textbf{sufficiently detected}.\par\smallskip The mutated response initially scores \textbf{4 out of 4} on item 4, including in the three additional scoring runs used to check whether the insufficient penalty persists. Under the refined item, it scores \textbf{0 out of 4}, while the reference response scores \textbf{4 out of 4}. The judge explains: ``Arithmetic is inconsistent: 6 billion times 78,000 equals 4.68e14, not 4.7e13 or 47 trillion.''}
\RMAppendixFull{\textbf{What this example shows.} The initial rubric already requires correct arithmetic, but the judge still gives full credit. The refinement makes two checks explicit: each intermediate calculation must be correct, and the final estimate must follow from those calculations.}
\bottomrule
\end{longtable}
\endgroup

\begingroup
\fontsize{9}{10.8}\selectfont
\setlength{\tabcolsep}{0pt}
\renewcommand{\arraystretch}{1.03}
\edef\RMAppendixFullWidth{\the\linewidth}
\edef\RMAppendixHalfWidth{\the\dimexpr(\linewidth-14pt)/2\relax}
\setlength{\LTcapwidth}{\linewidth}
\begin{longtable}{@{}>{\raggedright\arraybackslash}p{\RMAppendixHalfWidth}@{\hspace{14pt}}>{\raggedright\arraybackslash}p{\RMAppendixHalfWidth}@{}}
\caption{\textbf{Example 3: Checking that form submission displays the results.} Code, sample 577.}
\label{tab:refinement-example-form-submission}\\
\toprule
\endfirsthead
\multicolumn{2}{@{}l@{}}{\textit{Table~\thetable\ (continued). Example 3: Form submission.}}\\[4pt]
\toprule
\endhead
\midrule
\multicolumn{2}{r@{}}{\textit{Continued on the next page.}}\\
\endfoot
\endlastfoot
\RMAppendixFull{\textbf{Task.} ``i have a form with two input: job\_title, work\_city, and a button which can submit\par make a django controller accept these two input argument, and print it on result page''}
\RMAppendixFull{\textbf{Selected operator.} \textit{Break logic $\times$ Logical correctness.} Change the target of a function call so that it performs the wrong action. Here, the operator changes which page the code displays after the user submits the form.}
\RMAppendixFull{\textbf{Reading the code.} Django is a Python web framework. The \texttt{render} function displays the named page using the supplied values. In both responses, \texttt{job\_form.html} contains the input form, and \texttt{result.html} displays the submitted job title and city. The excerpts show the code after it has read these two values.}
\midrule
\textbf{Reference response $y^{\mathrm{ref}}$ (excerpt)}
&
\textbf{Mutated response $\tilde y$ (same code)}
\\*[5pt]
\texttt{return render(request,}\par
\hspace*{1em}\texttt{"\textbf{result.html}", \{}\par
\hspace*{1em}\texttt{"job\_title": job\_title,}\par
\hspace*{1em}\texttt{"work\_city": work\_city}\par
\texttt{\})}
&
\texttt{return render(request,}\par
\hspace*{1em}\texttt{"\textbf{job\_form.html}", \{}\par
\hspace*{1em}\texttt{"job\_title": job\_title,}\par
\hspace*{1em}\texttt{"work\_city": work\_city}\par
\texttt{\})}
\\[5pt]
\RMAppendixFull{\textbf{Injected defect.} The submitted values are read correctly, but the code returns the input form instead of the result page. That form has no code to display the submitted values, so the user sees empty input fields. The mutation changes the page name in the code and updates the corresponding sentence in the explanation.}
\midrule
\textbf{Relevant item in the initial rubric}
&
\textbf{The same item in the refined rubric}
\\*[5pt]
\textit{Item 2, weight 3.}\par ``The response provides an HTML template or return statement that correctly passes the two inputs to a result page.''
&
\textit{Item 2, weight 3 (unchanged).}\par ``The response provides a result-page response path that, \textbf{after a successful form submission}, passes the submitted `job\_title' and `work\_city' values to the result page \textbf{rather than returning the user to the input form or another unrelated page}.''
\\[5pt]
\RMAppendixFull{\textbf{What changed in the rubric.} Item 2 makes the required behavior explicit: submitting the form must display the results. Item 3 is also revised to check that the form and URL configuration use matching names, following a different injected defect. No items are added, and all six item weights remain unchanged.}
\midrule
\RMAppendixFull{\textbf{Scores before and after refinement.}\par\smallskip \textit{Initial rubric:} reference 100.00; mutated 100.00. The defect is \textbf{not sufficiently detected}.\par \textit{Refined rubric:} reference 100.00; mutated 88.46. The defect is \textbf{sufficiently detected}.\par\smallskip Although the initial item mentions a result page, the mutated response receives full credit on all six items. Under the refined item 2, it scores \textbf{2 out of 4}, while the reference response retains \textbf{4 out of 4}. All other item scores remain unchanged, so the 11.54-point overall decrease comes entirely from this item.}
\RMAppendixFull{\textbf{What this example shows.} A response can include the input form, the result page, and code that reads the inputs, yet still display the wrong page after submission. The mutation exposes this missed error, and refinement makes the rubric check the behavior of the submitted code more explicitly.}
\bottomrule
\end{longtable}
\endgroup

\endgroup

\section{Threshold Sensitivity}
\label{app:threshold-sensitivity}

Recall that \ours considers an injected defect sufficiently penalized if either the overall score decreases by at least $\delta_{\mathrm{total}}$ or some rubric-item score decreases by at least $\delta_{\mathrm{item}}$. We use $(\delta_{\mathrm{total}}, \delta_{\mathrm{item}})=(8,2)$ as the default setting.

The two thresholds operate at different levels of the rubric score. Each rubric item is scored on a 0--4 scale, whereas the weighted aggregate score is normalized to 0--100. Because each mutation is designed to introduce a targeted defect, its effect may be concentrated on one or a few rubric items and consequently be attenuated after aggregation. The item-level threshold therefore allows a clear local score decrease to count as sufficient even when the corresponding change in the aggregate score is modest. We set $\delta_{\mathrm{item}}=2$, corresponding to a two-point decrease on the 0--4 item scale, and use $\delta_{\mathrm{total}}=8$ as the corresponding aggregate-score threshold. Importantly, our method does not rely on these particular values being optimal.

We examine sensitivity to both thresholds on RM-Chat with Gemini-2.5-flash-lite using a one-factor-at-a-time analysis. Specifically, we vary $\delta_{\mathrm{total}}\in\{4,8,12,16\}$ while fixing $\delta_{\mathrm{item}}=2$, and vary $\delta_{\mathrm{item}}\in\{1,2,3,4\}$ while fixing $\delta_{\mathrm{total}}=8$. Because $(8,2)$ appears in both sweeps, these experiments cover seven unique threshold settings.

\begin{figure}[t]
    \centering
    \includegraphics[width=0.7\linewidth]{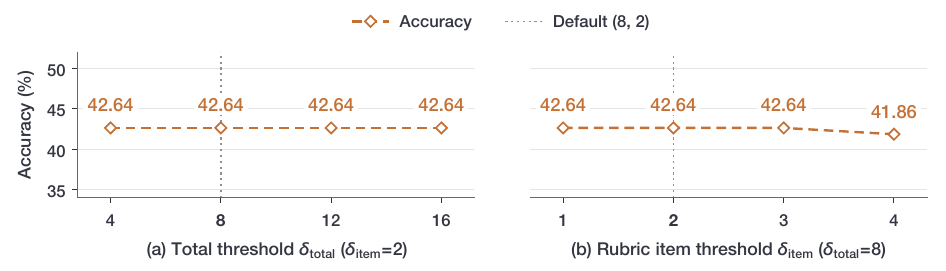}
    \captionof{figure}{Threshold sensitivity on RM-Chat with Gemini-2.5-flash-lite. Each panel varies one threshold while fixing the other at its default value. Final evaluation accuracy is shown on the same percentage scale. Dotted lines indicate the default setting $(\delta_{\mathrm{total}},\delta_{\mathrm{item}})=(8,2)$.}
    \label{fig:threshold-sensitivity}
\end{figure}

As shown in Figure~\ref{fig:threshold-sensitivity}, the final evaluation accuracy is highly stable across these settings. Six of the seven threshold combinations yield exactly the same accuracy of 42.64\%, including all four values of $\delta_{\mathrm{total}}$ when $\delta_{\mathrm{item}}=2$ and $\delta_{\mathrm{item}}\in\{1,2,3\}$ when $\delta_{\mathrm{total}}=8$. Only the most stringent item-level setting, $(\delta_{\mathrm{total}},\delta_{\mathrm{item}})=(8,4)$, produces a slightly lower accuracy of 41.86\%.

These results indicate that the effectiveness of \ours is not sensitive to a narrow choice of threshold values. We therefore use $(8,2)$ as a fixed default throughout the main experiments rather than tuning the thresholds separately for individual datasets or LLMs.

\end{document}